\documentclass[final]{cvpr}

\usepackage{times}
\usepackage{epsfig}
\usepackage{graphicx}
\usepackage{amsmath}
\usepackage{amssymb}

\usepackage{float}
\usepackage{booktabs}
\usepackage{multirow}
\usepackage{multicol}
\usepackage{wrapfig}
\usepackage{subcaption}
\usepackage{caption}
\usepackage[numbers,sort]{natbib}
\usepackage[dvipsnames]{xcolor}

\definecolor{preact}{HTML}{c994c7}
\definecolor{actact}{HTML}{dd1c77}
\definecolor{aqua}{HTML}{00FFFF}
\graphicspath{{./figures/}}

\usepackage[pagebackref=true,breaklinks=true,colorlinks,bookmarks=false]{hyperref}

\def\cvprPaperID{10} 
\def\confYear{CVPR 2021}

\begin{document}
\title{CoCon: Cooperative-Contrastive Learning}

\author{
  Nishant Rai\textsuperscript{1}, Ehsan Adeli\textsuperscript{1}, Kuan-Hui Lee\textsuperscript{2}, Adrien Gaidon\textsuperscript{2}, Juan Carlos Niebles\textsuperscript{1} \\
  \textsuperscript{1}Stanford University \quad \textsuperscript{2}Toyota Research Institute\\
}


\twocolumn[{%
\renewcommand\twocolumn[1][]{#1}%
\maketitle
\begin{center}
    \centering
    \includegraphics[width=0.8\linewidth]{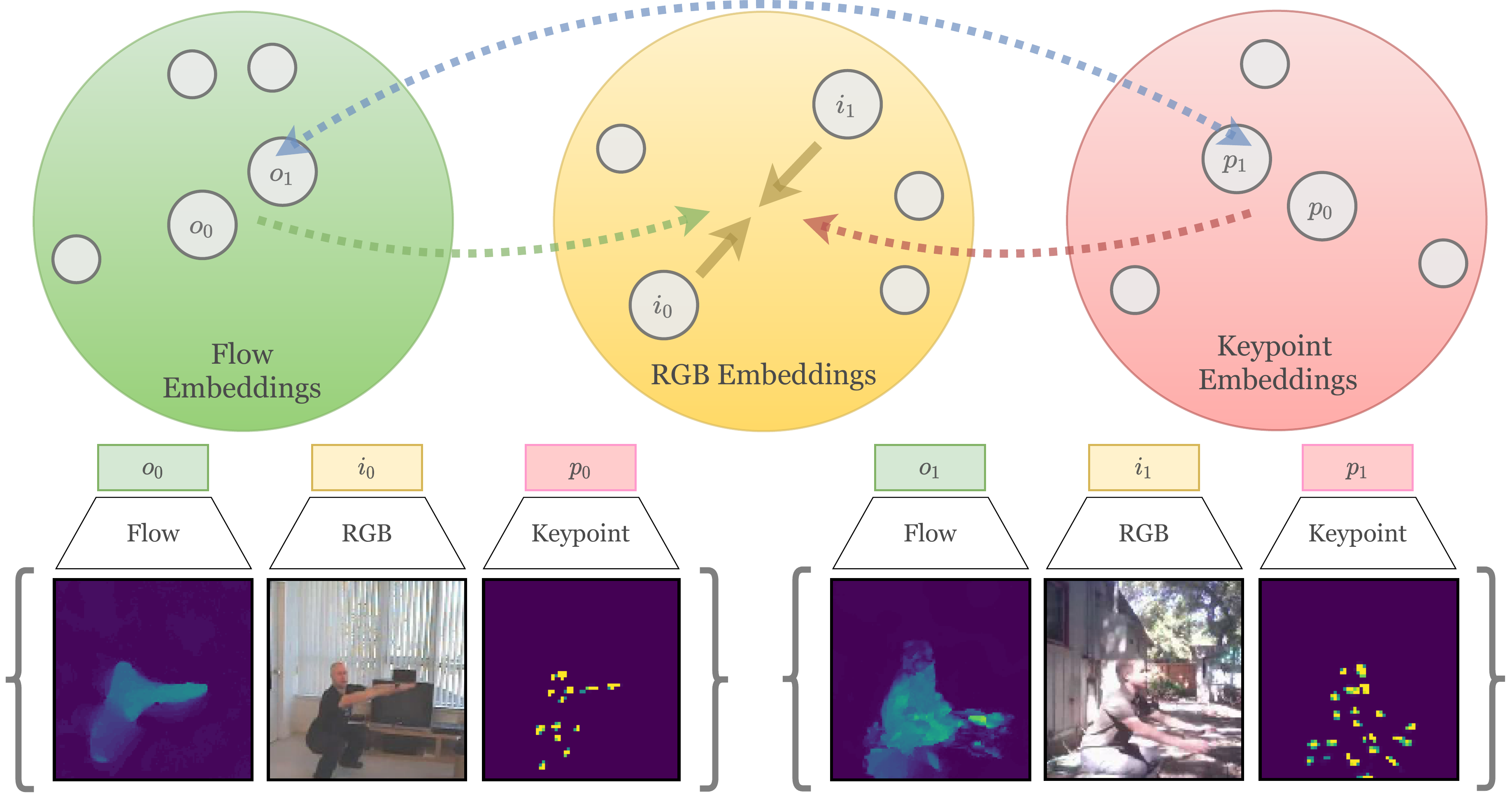}
    \captionof{figure}{\small{Given a pair of instances (e.g. people doing squats) and corresponding multiple views, features are computed using view-specific deep encoders $f$'s. Different instances may have contrasting similarities in different views. For instance, $V_0$ (left) and $V_1$ (right) have similar optical-flow $o = f_{flow}$ and pose keypoints (keypoint) $p = f_{keypoint}$ features but their image $i = f_{rgb}$ features are far apart. CoCon leverages these inconsistencies by encouraging the distances in all views to become similar. High similarity of $o_0$, $o_1$ and $p_0$, $p_1$ nudges $i_0$, $i_1$ towards each other in the RGB space.}}
    \label{intuition}
    \centering
\end{center}
}]

\begin{abstract}
Labeling videos at scale is impractical. Consequently, self-supervised visual representation learning is key for efficient video analysis. Recent success in learning image representations suggest contrastive learning is a promising framework to tackle this challenge. However, when applied to real-world videos, contrastive learning may unknowingly lead to separation of instances that contain semantically similar events. In our work, we introduce a cooperative variant of contrastive learning to utilize complementary information across views and address this issue. We use data-driven sampling to leverage implicit relationships between multiple input video views, whether observed (e.g. RGB) or inferred (e.g. flow, segmentation masks, poses). We are one of the firsts to explore exploiting inter-instance relationships to drive learning. We experimentally evaluate our representations on the downstream task of action recognition. Our method achieves competitive performance on standard benchmarks (UCF101, HMDB51, Kinetics400). Furthermore, qualitative experiments illustrate that our models can capture higher-order class relationships. The code is available at {\small\url{http://github.com/nishantrai18/CoCon}}.
\end{abstract}

\section{Introduction}

There has recently been a surge in interest for approaches utilizing self-supervised methods for visual representation learning. Recent advances in visual representation learning have demonstrated impressive performance compared to their supervised counterparts \cite{simclr, moco}. Fresh development in the video domain have attempted to make similar improvements \cite{lotter2016deep, crossLearn, hu2020probabilistic, dpc}.

Videos are a rich source for self-supervision, due to the inherent temporal consistency in neighboring frames. A natural approach to exploit this temporal structure is predicting future context as done in \cite{deepmsvid, lotter2016deep, hu2020probabilistic, dpc}. Such approaches perform future prediction in mainly two ways: (1) predicting a reconstruction of future frames \cite{deepmsvid, lotter2016deep, srivastava2015unsupervised}, (2) predicting features representing the future frames \cite{hu2020probabilistic, dpc}. If the goal is learning high-level semantic features for other downstream tasks, then complete reconstruction of frames is unnecessary. Inspired by developments in language modelling \cite{mnih2013learning}, recent work \cite{vondrickUnsupEmbVid} propose losses that only focus on the latent embedding using frame-level context. One of the more recent approaches \cite{dpc} propose utilizing spatio-temporal context to learn meaningful representations. Even though such developments have led to improved performance, the quality of the learned features is still lagging behind that of their supervised counterparts.

Due to the lack of labels in self-supervised settings, it is impossible to make direct associations between different training instances. Instead, prior work has learned associations based on structure, either in the form of temporal \cite{kim2019self, wei2018learning, lee2017unsupervised, lt_motion, dpc} or spatial proximity \cite{kim2019self, noroozi2016unsupervised, Jing2018SelfsupervisedSF, dpc} of patches extracted from training images or videos. However, the contrastive losses utilized enforce similarity constraints between instances from same videos while pushing instances from other videos far away even if they represent the same semantic content. This inherent drawback forces learning of features with limited semantic knowledge and encourage performing low-level discrimination between different videos. Recent approaches suffer from this restriction leading to poor representations.

The idea of utilizing multiple views of information has been a well-established one with roots in human perception \cite{prediction, predictive}. It's argued that useful higher order semantics are present throughout different views and are consistent across them. At the same time, different views provide complementary information which  can be utilized to aid learning in other views. Multi-view learning has been a popular direction \cite{cmc, crossLearn} utilizing these traits to improve representation quality. Recent approaches learn features utilizing multiple views with the motivation that information shared across views has valuable semantic meaning. A majority of these approaches directly utilize core ideas such as contrastive learning \cite{oord2018representation} and mutual information maximization \cite{li2018survey, arora2019theoretical, wu2018unsupervised}. Although the fusion of views leads to improved representations, such approaches also utilize contrastive losses, consequently suffering from the same drawback of low-level discrimination between similar instances.

We propose Cooperative Contrastive Learning (CoCon), which overcomes this shortcoming and leads to improved visual representations. Our main motivation is that each view sees a specific pattern, which can be useful to guide other views and improve representations. Our approach utilizes inter-view information to avoid the drawback of discriminating similar instances discussed earlier. To this end, each view sees a different aspect of the videos, allowing it to suggest potentially similar instances to other views. This allows us to infer implicit relationships between instances in a self-supervised multi-view setting, something which we are the first to explore. These associations are then used in order to learn better representations for downstream applications such as video classification and action recognition. Fig. \ref{intuition} shows an overview of CoCon. It is worth noting that although CoCon utilizes building blocks currently used in self-supervised representation learning, it is applicable to other tasks utilizing contrastive learning and be used in conjunction with other recently proposed methods.

We use `freely' available views of the input such as RGB frames and Optical Flow. We also explore the benefit of using high-level inferred semantics as additional noisy views, such as human pose keypoints and segmentation masks generated using off-the-shelf models \cite{detectron2}. These views are not independent, as they can be derived from the original input images. However, they are complementary and lead to significant gains, demonstrating CoCon's effectiveness even with noisy related views. The extensible nature of our framework and the `freely' available views used make it possible to use CoCon with any publicly available video dataset and other contrastive learning approaches.

\section{Related Work}

\textbf{Self-supervised Learning from images} Recent approaches have tackled image representation learning by exploiting color information \cite{zhang2016colorful, larsson2017colorization} and spatial relationships \cite{noroozi2016unsupervised,sabokrou2019self}, where relative positions between image patches are exploited as supervisory signals. Several approaches apply self-supervision to super-resolution \cite{dong2014learning,johnson2016perceptual} or even to multi-task \cite{doersch2017multi} and cross-domain \cite{ren2018cross} learning frameworks.

\textbf{Self-supervised Learning from videos} Multiple approaches \cite{deepmsvid, lotter2016deep, hu2020probabilistic, srivastava2015unsupervised, dpc} perform self-supervision through `predicting' future frames. However, the term `predicting' is overloaded, as they do not directly predict and reconstruct frames but instead operate on latent representations. This ignores stochasticity of frame appearance, e.g., illumination changes, camera motion, appearance changes due to reflections and so on, allowing the model to focus on higher-order semantic features. Recent work \cite{cmc, dpc} utilize Noise Contrastive Estimation to perform prediction of the latent representations rather than the exact future frames, vastly improving performance. Yet, another class of proxy tasks are based on temporal ordering of frames \cite{misra2016shuffle, wei2018learning}. Temporal coherence \cite{jayaraman2016slow, wang2015unsupervised} and 3D puzzle \cite{kim2019self} were used as proxy loss to exploit spatio/temporal structures.

\textbf{Multi-view learning} Multiple views of videos are rich sources of information for self-supervised learning \cite{crossLearn, cmc, vgan}. Two stream networks for action recognition \cite{simonyanTwoStream14} have led to many competitive approaches, which demonstrate using even derivable views such as optical flow helps improve performance considerably. There have been approaches \cite{lt_motion, vgan, cmc, crossLearn} utilizing diverse views, sometimes derivable from one other, to learn better representations. However, these approaches utilize inter-view links by maximizing mutual information between them.  Although this leads to improved performance, we believe the rich inter-view linkages can be utilized more effectively by utilizing them to uncover implicit relationships between instances.

\textbf{Multi-View Self-supervised learning} Multiple recent approaches \cite{Han20, han2020self, xdc, gdt} have tackled the challenge of multi-modal self-supervised learning achieving impressive performance. However, these approaches suffer from the same drawback of discriminating between similar instances, leaving potential to benefit from inter-sample relationships.

Most approaches above perform self-supervision using positive and negative pairs mined through structural constraints, e.g., temporal and spatial proximity. Although this results in representations that capture some degree of semantic information, it incorrectly leads to treating similar actions differently due to the inherent nature of their pair-mining. For instance, clip pairs in different videos are considered negatives, even if they represent the same action. We argue that utilizing different views and inter-instance relationships to propose positive pairs to aid training can lead to improvement of all views simultaneously.

\section{Method}

\begin{figure*}[t]
    \centering
    \begin{minipage}{0.45\linewidth}
        \centering
        \includegraphics[width=\linewidth]{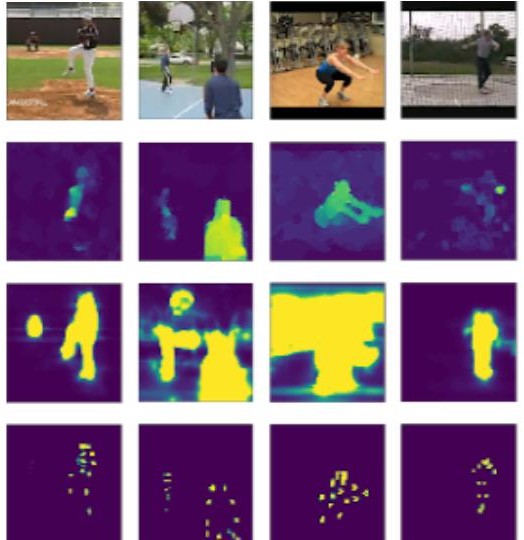}
        \captionof{figure}{\small{Examples for each view. From top to bottom - RGB, Flow, SegMasks and Poses. Note the prevalence of noise in a few samples, specially SegMasks; There are multiple other instances where Poses, SegMasks are noisy but have not been shown here.}}
        \label{views-samples}
    \end{minipage}~~
    \begin{minipage}{0.53\linewidth}
          \centering
          \renewcommand{\arraystretch}{1.0}
          {
          \small
            \begin{tabular}{c|c|c|c|c|c}
                \toprule
                View & Random & $\mathcal{L}_{cpc}$ & $\mathcal{L}_{sim}^{cpc}$ & $\mathcal{L}_{sync}^{cpc}$ &  $\mathcal{L}_{cocon}$\\
                \midrule
                RGB    & 46.7 & 63.7 & 66.0 & 62.7 & \textbf{67.8} \\
                Flow   & 65.3 & 69.8 & 71.4 & 69.2 & \textbf{72.5} \\
                \bottomrule
            \end{tabular}
            \captionof{table}{\small{Impact of losses on performance of models when jointly trained with RGB and Flow. CoCon i.e. $\mathcal{L}_{total}$ (67.8) comfortably improves performance over CPC i.e. $\mathcal{L}_{cpc}$ (63.7). $\mathcal{L}_{y}^{x} = \mathcal{L}_x + \lambda \mathcal{L}_y$ where $\lambda=10.0$ for this experiment}}
            \label{table:loss_exp}
          }
          \vspace{1mm}
          {\small
            \renewcommand{\arraystretch}{1.0}
            \begin{tabular}{c|c|c|c|c|c}
                \toprule
                \multirow{2}{*}{Method} & \multirow{2}{*}{Pretrain} & \multicolumn{2}{c|}{RGB} & \multicolumn{2}{c}{Flow\footnotemark[2]} \\
                 & & \multicolumn{1}{c}{UCF} & HMDB & \multicolumn{1}{c}{UCF} & HMDB \\
                \midrule
                {Random} &        & 46.7 & 20.6 & 65.3 & 31.2 \\
                {CPC}    & {K400} & 68.6 & 35.5 & 69.8 & 40.8\\
                {CoCon}   & {UCF}  & 67.8 & 37.7 & \textbf{72.5}  & 44.1 \\
                {CoCon}   & {K400} & \textbf{72.1} & \textbf{46.5} & 71.8 & \textbf{44.2} \\
                \bottomrule
            \end{tabular}
            \captionof{table}{\small{Impact of pre-training comparison. CoCon demonstrates a consistent improvement in both RGB and Flow.}}
            \label{table:dataset_size}
          }
          \vspace{1mm}
          {\small
            \centering
            \renewcommand{\arraystretch}{1.0}
            \renewcommand{\tabcolsep}{1mm}
            \begin{tabular}{c|c|c|c|c|c|c|c|c}
                \toprule
                \multirow{2}{*}{Method} & \multicolumn{2}{c|}{RGB} & \multicolumn{2}{c|}{Flow} & \multicolumn{2}{c|}{PoseHM} & \multicolumn{2}{c}{SegMask} \\
                 & \multicolumn{1}{c}{UCF} & HMDB & \multicolumn{1}{c}{UCF} & HMDB & \multicolumn{1}{c}{UCF} & HMDB & \multicolumn{1}{c}{UCF} & HMDB \\
                \midrule
                Random   & 46.7 & 20.6  & 65.3 &  31.2 & 51.7 & 33.0 & 42.7 & 26.3 \\
                CPC      & 63.7 & 33.1 & 71.2 &  44.6 & 56.4 & 42.0 & 53.7 & 32.8 \\
                CoCon     & \textbf{71.0} & \textbf{39.0} & \textbf{74.5} & \textbf{45.4} & \textbf{58.7} & \textbf{42.6} & \textbf{55.8} & \textbf{34.0} \\
                \bottomrule
            \end{tabular}
            \captionof{table}{\small{Impact of co-training on views. CoCon is jointly trained with four modalities (RGB, Flow, PoseHM, \& SegMask).}}
            \label{table:modes}
           }
    \end{minipage}
\end{figure*}

We describe cooperative contrastive learning (CoCon) and intuition behind our designs in this section. Additional details regarding architecture and implementation are present in the appendix. In the following sections, we build our framework borrowing the learning framework present in \cite{dpc} which learns video representations through spatio-temporal contrastive losses. It should be noted that even though we use this particular self-supervised backbone in our experiments, our approach is not restricted by the choice of the underlying self-supervised task. CoCon can be used in conjunction with any other frameworks currently present and allow them to be extended to a multi-view setting.

A video $V$ is a sequence of $T$ frames (not necessarily RGB images) with resolution $H \times W$ and $C$ channels, $\{\mathbf{i}_1, \mathbf{i}_2, \ldots, \mathbf{i}_{T}\}$, where $\mathbf{i}_t \in \mathbb{R}^{H \times W \times C}$. Assume $T = N * K$, where $N$ is the number of blocks and $K$ denotes the number of frames per block. We partition a video clip $V$ into $N$ disjoint blocks $V = \{\mathbf{x}_1, \mathbf{x}_2, \ldots, \mathbf{x}_{N}\}$, where $\mathbf{x}_j \in \mathbb{R}^{K \times H \times W \times C}$ and a non-linear encoder $f(.)$ transforms each input block $x_j$ into its latent representation $z_j = f(x_j)$. An aggregation function, $g(.)$ takes a sequence $\{z_1, z_2, \ldots, z_j\}$ as input and generates a context representation $c_j = g(z_1, z_2, \ldots, z_j)$. In our setup, $z_j \in \mathbb{R}^{H' \times W' \times D}$ and $c_j \in \mathbb{R}^{D}$. $D$ represents the embedding size and $H'$, $W'$ represent down-sampled resolutions as different regions in $z_j$ represent features for different spatial locations. We define $z'_j = Pool(z_j)$ where $z'_j \in \mathbb{R}^{D}$ and $c = F(V)$ where $F(.) = g(f(.))$.


Similar to \cite{dpc}, we create a prediction task involving predicting $z$ of future blocks. Details are provided in the appendix. For multiple views, we define $c_{v} = F_v(V_v)$, where $V_v$, $c_v$ and $F_v$ represent the input, context feature and composite encoder for view $v$ respectively.

\noindent\textbf{Contrastive Loss}
Noise Contrastive Estimation (NCE) \cite{gutmann2010noise, mnih2013learning, oord2018representation} constructs a binary classification task where a classifier is fed with real and noisy samples with the training objective being distinguishing them.
Similar to \cite{dpc, oord2018representation}, we use an NCE loss over our feature embeddings described in Eq \ref{lcpc}. $z_{i, k}$ represents the feature embedding for the $i^{th}$ time-step and the $k^{th}$ spatial location. Recall ${z}_j \in \mathbb{R}^{H' \times W' \times D}$ which preserves the spatial layout. We normalize $z_{i, k}$ to lie on the unit hypersphere. Eq \ref{lcpc} is a cross-entropy loss distinguishing one positive pair from all the negative pairs present in a video. We use temperature $\tau = 0.005$ in our experiments. In a batch setting with multiple video clips, it is possible to have more inter-clip negative pairs.  

To extend this to multiple views, we utilize different encoders $\phi_v$ for each view $v$. We train these encoders by utilizing $\mathcal{L}_{cpc}$ for each of them independently, giving us, $\mathcal{L}_{cpc} = \sum_{v} \mathcal{L}_{cpc}^{v}$

\begin{equation}
    \label{lcpc}
    \mathcal{L}_{cpc} = - \sum_{i, k} \bigg( \text{log} \frac{\text{exp}(\Tilde{z}_{i, k} \cdot z_{i, k}  \mathbin{/} \tau)}{\sum_{j, m} \text{exp}(\Tilde{z}_{i, k} \cdot z_{j, m} \mathbin{/} \tau)} \bigg)
\end{equation}

\noindent\textbf{Cooperative Multi-View Learning}
Recent approaches \cite{cmc, han2020self, crossLearn} tackle multi-view self-supervised learning by maximizing mutual information across views. They involve using positive and negative pairs generated using structural constraints, e.g., spatio-temporal proximity in videos \cite{dpc, Han20, crossLearn, cmc}. Although such representations capture semantic content, they unintentionally encourage discriminating video clips containing semantically similar content due to the inherent nature of pair generation, i.e. video clips from different videos are negatives. We utilize inter-instance relationships to alleviate some of these issues.

We soften this constraint by indirectly deriving pair proposals using different views. Such a co-operative scheme benefits all models as each individual view gradually improves. Better models are able to generate better proposals, improving performance of all views creating a positive feedback loop. Our belief is that significant semantic features should be universal across views, therefore, potential incorrect proposals from one view should cancel out through proposals from other views.

We achieve the above by computing view-specific distances and synchronizing them across all views. We enforce a consistency loss between distances from each view. Looking at it from another perspective, we are encouraging relationships between instances to be the same across views i.e. similar pairs in one view should be a similar pair in other views as well. Treating this as inter-view graph regularization, we create a graph similarity matrix $W_{v}$ of size $K \times K$, using some distance metric. We represent our distance metric by $\mathcal{D}(.)$. In our experiments, we use the cosine distance which translates to $W^{v}_{ab} = z_z \cdot z_b$.

Assume $h_{v}^a$ denotes the representation for the $v^{th}$ view of instance $a$. In our experiments, we use $h = z'$ giving us block level features. Our resultant loss becomes the inconsistency between similarity matrices across views. The resultant graph regularization loss becomes $\sum_{v_0, v_1} \| W^{v_0} - W^{v_1}\|$ which is simplified in Eq \ref{eq4}.

Building on top of our earlier intuition, in order to have sensible proposals, we need to have discriminative scores, i.e. we should have both positive ($\mathcal{D} \to 0$) and negative ($\mathcal{D} \to 1$) pairs. To promote well distributed distances, we utilize the hinge loss described in Eq \ref{eq5}.

$\mathcal{L}_{sim}$ is the hinge loss, where the first term pushes representations of the same instance in different views closer; while the second term pushes different instances apart. Since the number of structural negative pairs are much larger than the positives, we introduce $\mu$ in order to balance the loss weights. We choose $\mu$ such that the first and second components contribute equally to the loss.

\begin{equation}
    \label{eq4}
    \mathcal{L}_{sync} = \sum_{v_0, v_1} \sum_{a, b} \Big( \mathcal{D}(h_{v_0}^a, h_{v_0}^{b}) - \mathcal{D}(h_{v_1}^{a}, h_{v_1}^{b}) \Big)^2
\end{equation}
\vspace{-3mm}
\begin{equation}
    \label{eq5}
    \begin{split}
    \mathcal{L}_{sim} = {} & \sum_{v_0, v_1} \sum_{a} \mathcal{D}(h_{v_0}^{a}, h_{v_1}^{a}) \\
    & + \mu \sum_{a \neq b} max \big( 0, 1 - \mathcal{D}(h_{v_0}^{a}, h_{v_1}^{b}) \big)
    \end{split}
\end{equation}

Note that $\mathcal{L}_{sim}$ entangles different views together. An alternative would be defining such a loss individually for each view. However, diversity is inherently encouraged through $\mathcal{L}_{cpc}$, and interactions between views have the side-effect of increasing their mutual information (MI), which leads to improved performance \cite{cmc, crossLearn}.

We combine the above losses to get our cooperative loss, $\mathcal{L}_{coop} = \mathcal{L}_{sync} + \alpha \cdot \mathcal{L}_{sim}$. We use $\alpha = 1.0$ for our experiments and observe roughly similar performance for different values of $\alpha$. The overall loss of our model is given by $\mathcal{L}_{cocon} = \mathcal{L}_{cpc} + \lambda \cdot \mathcal{L}_{coop}$. $\mathcal{L}_{cpc}$ encourages our model to learn good features for each view, while $\mathcal{L}_{coop}$ nudges it to learn higher-level features using all views while respecting the similarity structure across them.

\section{Experiments}

\begin{table*}
    \renewcommand{\arraystretch}{1.0}
    \renewcommand{\tabcolsep}{1mm}
    {\small
    \begin{minipage}{.67\linewidth}
      \centering
    {\small
        \begin{tabular}{c|c|c}
            \toprule
            {Action Class} & CoCon & CPC\\
            \midrule
                PlayCello & PlaySitar, PlayTabla, PlayDhol &  N/A \\
                Skiing & Surfing, Skijet & Surfing \\
                HammerThrow & BaseballPitch, ThrowDiscus, Shotput &  N/A\\
                BrushTeeth & ApplyLipstick, EyeMakeup, ShaveBeard & ApplyLipstick\\
            \bottomrule
        \end{tabular}
        }
        \captionof{table}{\small{Nearest consistent semantic classes. Individually trained views (\textit{CPC}) do not have consistent neighbors across views, leading to empty results (N/A) for 'PlayingCello' and 'HammerThrow'. While views trained using CoCon show consistency across views, leading to sensible relationships e.g. 'HammerThrow' related to other classes involving throwing.}}
        \label{table:nn_classes}
    \end{minipage}%
    ~~~
    \begin{minipage}{.32\linewidth}
      \centering
            \begin{tabular}{c|c|c|c|c}
                \toprule
                \multirow{2}{*}{\# Views} & \multicolumn{2}{c|}{RGB} & \multicolumn{2}{c}{Flow} \\
                 & \multicolumn{1}{c}{UCF} & HMDB & \multicolumn{1}{c}{UCF} & HMDB \\
                \midrule
                2 & 67.8 & 37.7  & 72.5 & 44.1 \\
                4 & \textbf{71.0} & \textbf{39.0} & \textbf{74.5} & \textbf{45.4} \\
                \bottomrule
            \end{tabular}
             \captionof{table}{\small{Impact of performance on varying views. A consistent improvement can be seen with more views despite the prevalent noise in PoseHM and SegMasks.}}
             \label{table:coop}
    \end{minipage}
    }
\end{table*}

The goal of our framework is to learn video representations which can be leveraged for video analysis tasks. Therefore, we perform experiments validating the quality of our representations. We measure downstream action classification to objectively measure model effectiveness and analyze impact of our designs through controlled ablation studies. We also conduct qualitative experiments to gain deeper insights into our approach. In this section, we briefly go over our experiment framework. Additional details and discussions for each component are provided in the appendix.

\textbf{Datasets} Our approach is a self-supervised learning framework for any dataset with multiple views. However, we discuss its relevance to video action classification in our experiments. We focus on human action datasets i.e. UCF101, HMDB51 and Kinetics400. UCF101 contains 13K videos spanning over 101 human action classes. HMDB51 contains 7K video clips mostly from movies for 51 classes. Kinetics-400 (K400) is a large video dataset with 306K video clips from 400 classes.

\textbf{Views} We utilize different views in our experiments. For Kinetics-400, we learn encoders for RGB and Optical Flow. We use Farneback flow (FF) \cite{farneback} instead of the commonly used TVL1-Flow as it is quicker to compute lowering our computation budget. Although FF leads to lower performance compared to TVL1, the essence of our claims remain unaffected. For UCF101 and HMDB51, we learn encoders for RGB, TVL1 Optical Flow, Pose Heatmaps (PoseHMs) and Human Segmentation Masks (SegMasks). A few visual samples for each view are provided in \ref{views-samples}. PoseHMs and SegMasks are generated using an off-the-shelf detector \cite{detectron2} without any form of pre/post-processing.

\label{impl}
\textbf{Implementation Details} We choose a 3D-ResNet similar to \cite{dpc, hara2018can} as the encoder $f(.)$. We choose $N=8$ and $K=5$ in our experiments. We subsample the input by uniformly choosing one out of every 3 frames. Our predictive task involves predicting the last three blocks using the first five blocks. We use standard data augmentations during training whose details are provided in the appendix. We train our models using Adam \cite{adam} optimizer with an initial learning rate of $10^{-3}$, decreased upon loss plateauing. We use 4 GPUs with a batch size of 16 samples per GPU. Multiple spatio-temporal samples ensure sufficient negative examples despite the small batch size used for training.

\textbf{Action Classification} We measure the effectiveness of our learned representations using the downstream task of action classification. We follow the standard evaluation protocol of using self-supervised model weights as initialization for supervised learning. The architecture is then fine-tuned end-to-end using class label supervision. We finally report the fine-tuned accuracies on UCF101 and HMDB51. While fine-tuning, we use the learned composite function $F(.)$ in order to generate context representations for the video blocks. The context feature is further passed through a spatial pooling layer followed by a fully-connected layer and a multi-way softmax for action classification.

\subsection{Quantitative Results}

We analyze various aspects of CoCon through ablation studies, experiments on multiple datasets, controlled variation of views and comparison to comparable methods. We objectively evaluate model performance using downstream classification accuracy as a proxy for learned representation quality. Pre-training is performed on either UCF101 or Kinetics400. We propose two baselines for comparison. (1) \textit{Random} - random initialization of weights (2) \textit{CPC} - self-supervised training utilizing only $\mathcal{L}_{cpc}$; which is effectively individual training of views. \textit{CPC} serves as a critical baseline to measure the benefits of multi-view training as opposed to individual training.

\textbf{Ablation Study}
We have motivated the utility of our various loss components. We now perform experiments to quantify the impact of each. The pre-training dataset used is the $1^{st}$ split of UCF101, and downstream classification accuracy is computed on the same. Table \ref{table:loss_exp} summarizes the results of our experiment. As expected, all cross-view approaches comfortably perform better than \textit{CPC}; demonstrating the utility of multi-view training.

Using $\mathcal{L}_{sync}^{cpc}$ leads to no performance improvements, as only using $\mathcal{L}_{sync}$ leads to the model collapsing by squashing all $\mathcal{D}$ scores to have similar values, thus necessitating $\mathcal{L}_{sim}$ to counter-balance this tendency. $\mathcal{L}_{sim}^{cpc}$ leads to improved performance wrt $\mathcal{L}_{cpc}$ as it learns better features by effectively maximizing mutual information between views. CoCon i.e $\mathcal{L}_{cocon}$ achieves the same by also regularizing manifolds across views, leading to even better performance across all views. The important comparison to observe is between $\mathcal{L}_{sim}^{cpc}$ and $L_{cocon}$. As $\mathcal{L}_{sim}^{cpc}$ is the most similar baseline to other multi-view approaches, e.g., CMC \cite{cmc}. However, we argue this baseline is even stronger as it involves both single-view and multi-view components compared to \cite{cmc}, which only uses a contrastive multi-view loss to learn representations.

\begin{table*}
    \renewcommand{\arraystretch}{1.0}
    \centering
    {\small
    \begin{tabular}{lcccccc}
        \toprule
        \textbf{Method} & \textbf{Resolution} & \textbf{Backbone} & \textbf{\# Views} & \textbf{Pre-train} & \textbf{UCF101} & \textbf{HMDB51} \\
        \toprule
        Random Initialization & $128 \times 128$ & ResNet18 & 1 & & 46.7 & 20.6 \\
        ImageNet \cite{simonyan2014two} & $224 \times 224$ & VGG-M-2048 & 1 & ImageNet & 73.0 & 40.5 \\
        \midrule
        Shuffle and Learn \cite{misra2016shuffle}  & $227 \times 227$ & CaffeNet  & 1 & UCF-HMDB & 50.2 & 18.1 \\
        OPN \cite{lee2017unsupervised}             & $80 \times 80$ & VGG-M-2048 & 1 & UCF-HMDB & 59.8 & 23.8 \\
        DPC \cite{dpc}             & $128 \times 128$ & ResNet18  & 1 & UCF101   & 60.6 & - \\
        VGAN \cite{vgan}             & N/A & C3D                   &  2  & Flickr \cite{vgan} & 52.1 & - \\
        LT-Motion \cite{lt_motion}   & N/A & RNN \cite{lt_motion}  &  2  & NTU                & 53.0 & - \\
        Cross and Learn \cite{crossLearn} & $224 \times 224$ & CaffeNet     &  2  & UCF101 & 58.7 & 27.2 \\
        Geometry \cite{geometry}      & N/A & CaffeNet     &  2  & UCF101 & 55.1 & 23.3 \\
        CMC \cite{cmc}   & $128 \times 128$ & CaffeNet     &  3  & UCF101 & 59.7 & 26.1 \\
        CoCon - RGB       & $128 \times 128$ & ResNet18  & 4 & UCF101 & \textbf{70.5} & \textbf{38.4} \\
        CoCon - Ensemble & $128 \times 128$ & ResNet18  & 4 & UCF101 & \textbf{82.4} & \textbf{52.0} \\
        \midrule
        3D-RotNet \cite{Jing2018SelfsupervisedSF} & $112 \times 112$ & ResNet18  &  1 & Kinetics & 62.9 & 33.7 \\
        DPC \cite{dpc} & $128 \times 128$ & ResNet18 & 1 & Kinetics & 68.2 & 34.5 \\
        CoCon - RGB & $128 \times 128$ & ResNet18  & 2 & Kinetics & \textbf{71.6} & \textbf{46.0} \\
        CoCon - Ensemble & $128 \times 128$ & ResNet18  & 2 & Kinetics & \textbf{78.1} & \textbf{52.0} \\
        \midrule
        ST-Puzzle \cite{kim2019self} & $224 \times 224$ & ResNet18 & 1 &  Kinetics & 65.8 & 33.7 \\
        DPC \cite{dpc} & $224 \times 224$ & ResNet34 & 1 & Kinetics & 75.7 & 35.7 \\
        CoCon - RGB & $224 \times 224$ & ResNet34  & 2 & Kinetics & \textbf{79.1} & \textbf{48.5} \\
        CoCon - Ensemble  & $224 \times 224$ & ResNet34  & 2 & Kinetics & \textbf{82.0} & \textbf{53.1} \\
        \bottomrule
    \end{tabular}
    }
    \caption{\small{Comparison of classification accuracies on UCF101 and HMDB51, averaged over all splits.}}
    \label{table:sota}
\end{table*}

\textbf{Effect of Datasets}
A critical benefit of self-supervised approaches is the ability to run on large unlabelled datasets. To simulate such a setting, we perform pre-training using UCF101 or Kinetics400 \footnote{\label{ff_flow} Optical Flow used for Kinetics400 is Farneback Flow; as opposed to TVL1 Flow for UCF101 and HMDB51. This difference in pre-training and fine-tuning modalities leads to less than expected performance gains.} without labels utilizing the $1^{st}$ splits of UCF101 and HMDB51 for evaluation. Table \ref{table:dataset_size} confirms pre-training with a larger dataset leads to better performance. It is also worth noting that CoCon pre-trained with UCF101 outperforms \textit{CPC} trained on Kinetics400 even though CoCon on UCF101 uses only around $10\%$ data compared to Kinetics. Further demonstrating the potential of utilizing multiple views as opposed to training with larger and diverse datasets.

\begin{table*}
    \renewcommand{\arraystretch}{1.1}
    \renewcommand{\tabcolsep}{1mm}
    {
        \begin{minipage}{.68\linewidth}
        \centering
        \includegraphics[width=\linewidth]{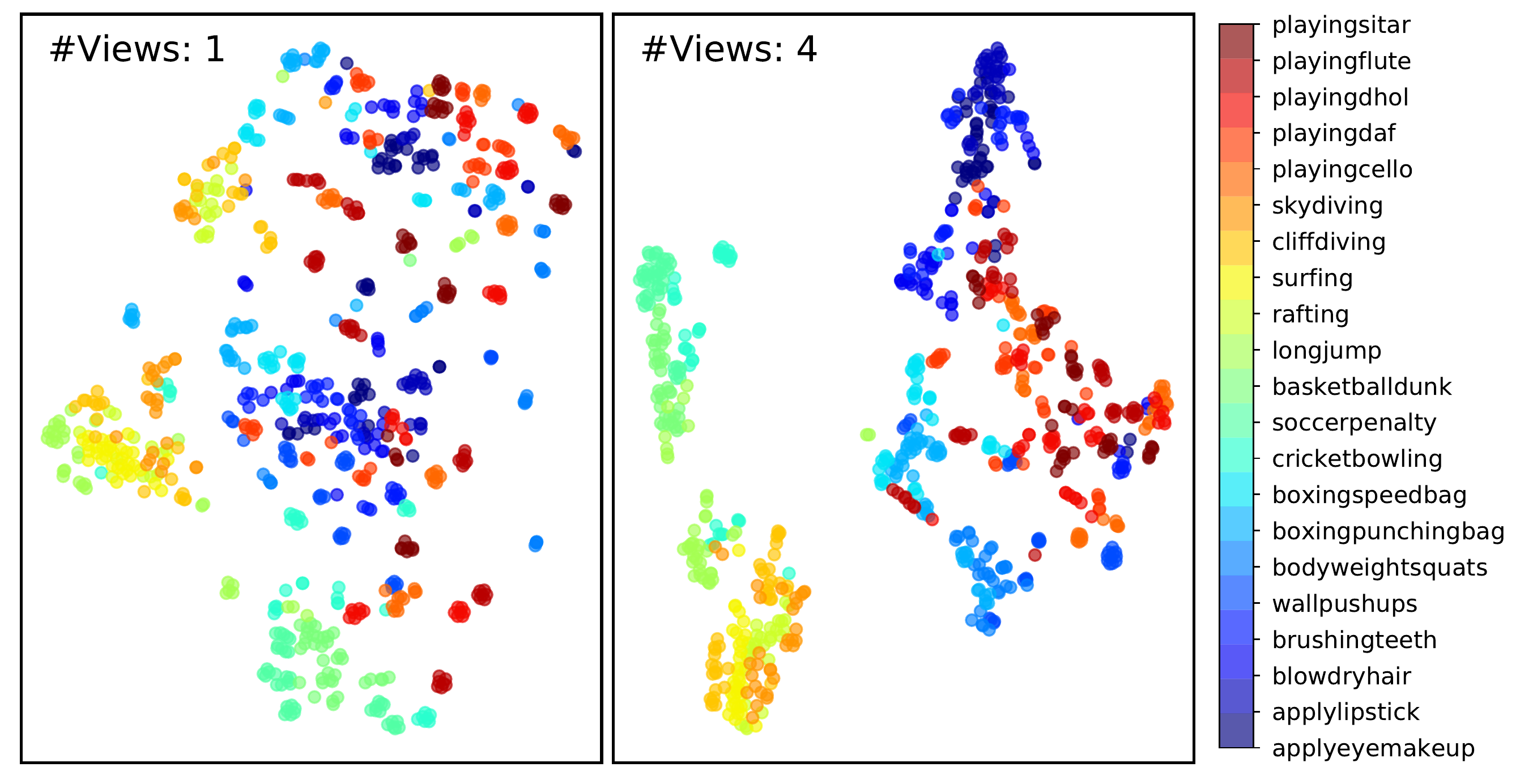}
        \captionof{figure}{\small{t-SNE visualization of RGB features from \textit{CPC} (left) and CoCon (right) trained with 4 views. The color mapping for each category represents the relationships between action classes, e.g., Red: Instruments; Yellow: Water Sports; Light-blue: Physical Acts; Blue: Makeup-Hygiene. More meaningful clusters are formed using CoCon; signifying the ability of CoCon to align different yet semantically-related classes without any additional supervision.}}
        \label{tsne}
    \end{minipage}
    } ~~
    \begin{minipage}{.3\linewidth}
        \centering
        \includegraphics[width=0.98\linewidth]{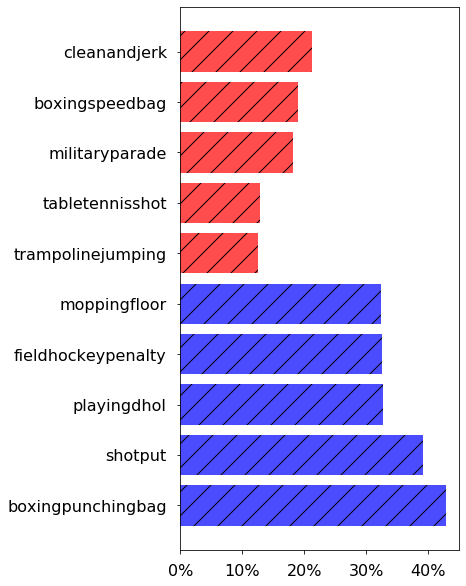}
        \captionof{figure}{\small{Differences between class-wise accuracy for CoCon vs CPC. Only extreme classes are displayed. \textcolor{blue}{Blue} - Gains; \textcolor{red}{Red} - Loss}}
        \label{class_acc}
    \end{minipage}
\end{table*}

When comparing the \textit{Random} baseline and CoCon pre-trained on Kinetics400, we observe higher performance gains for RGB (+25.4\%) compared to Optical-Flow (+6.9\%). We argue this is due to higher variance and complexity of RGB compared to Flow, allowing a randomly initialized network to perform relatively better with Flow. While comparing our approach with \textit{CPC}, we again observe higher gains in RGB (+4.1\%) compared to Flow (+2.7\%). This can be explained by the potential capability of RGB to capture flow-like features when learned jointly.

\textbf{Effect of cooperative training}
We compare benefits of cooperative training with varying views. We look at co-training of RGB, Flow, SegMasks and PoseHMs. Recall that these additional views are generated using off-the-shelf models without any additional post-processing. Even though they are somewhat redundant i.e. Flow, PoseHM, SegMasks are actually derived from RGB Images; using them simultaneously still leads to a large performance increase. We also note that although SegMasks and PoseHMs are sparse low-dimensional features, they still help improve performance across all views.

Table \ref{table:modes} summarizes downstream action recognition performance of each view under different approaches. We see improved performance with increase in the number of views used. Consistent gains for views such as Flow, SegMasks, PoseHM, which are not as expressive as RGB points towards extraction of higher-order features even from low dimensional inputs. We observe PoseHM and SegMask have lower performance gains when evaluated on HMDB51. This can be attributed to the large degree of noise in PoseHMs and SegMasks for HMDB51. HMDB is a challenging and diverse dataset, leading to poor predictions from our off-the-shelf detector. In conclusion, the benefits of joint training are apparent as CoCon leads to a performance improvement for all the views involved.

\textbf{Effect of additional views}
CoCon hinges on the assumption that multi-view information helps in improving overall representation quality. To verify our hypothesis, we study co-training with different number of views. We consider two scenarios, 1) Joint training of RGB and Flow streams, and 2) Joint training of RGB, Flow, SegMasks and PoseHMs. Table \ref{table:coop} shows a consistent increase across views when increasing the number of views used during training. We should note that both SegMasks and PoseHMs contain significant noise as the off-the-shelf models incorrectly detects and misses humans in numerous videos. However, we see a consistent mutual increase in performance for all the involved views despite the prevalence of noise.

\textbf{Comparison with comparable approaches}
We summarize comparisons of CoCon with comparable state-of-the-art approaches in Table \ref{table:sota}. CoCon-Ensemble refers to an ensemble of models for all the involved views. We observe a few major trends, (1) When pre-training on UCF101, using multiple views allows us to outperform the nearest comparable approach by around 10.4\%. This demonstrates the potential of cooperatively utilizing multiple views to learn representations. (2) We see considerable gains while training on Kinetics400 as well, however, the increase is smaller compared to UCF101. We argue the reasons are, a) we only utilize two views for co-training. b) the flow we utilize for Kinetics400 is Farneback Flow instead of TVL1 flow used for UCF101 and HMDB51. (3) Our method comfortably outperforms recent multi-view approaches consistently on UCF101 and HMDB51. (4) An interesting observation is that using multiple views of a small dataset (UCF101) performs better (71.0\%) than pre-training on a large dataset, Kinetics400 (68.2\%). This suggests that utilizing different views can be better than merely training on larger datasets.

\textbf{Comparison with recent approaches}
A few very recent approaches \cite{Han20, han2020self, xdc, gdt} have tackled multi-modal self-supervised achieving impressive performance. CoCon differs from them as it considers inter-instance relationships to aid learning in addition to relationships between views. Due to resource constraints, it was not possible to have a fair comparison due to the significant difference in the amount of GPUs, number of epochs trained and the backbones used. However, we hope our carefully constructed experiments given earlier provide deeper insights into CoCon's benefits even with lower resource requirements.

\begin{figure*}[h]
  \centering
    \includegraphics[width=0.87\linewidth]{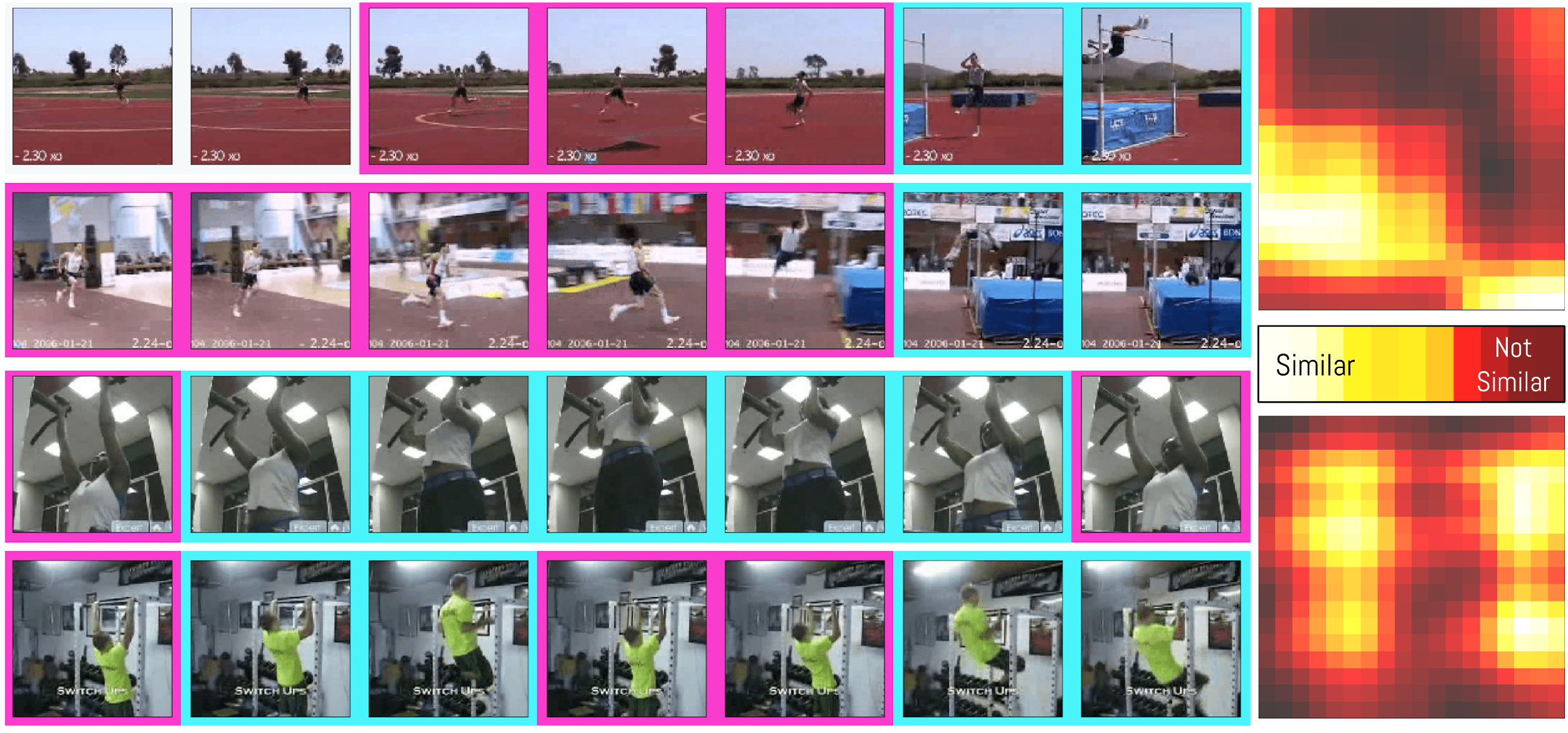}
  \caption{\small{Soft Alignment of videos from UCF101 test split using CoCon pre-trained on UCF101. The first pair of videos involves pull-ups; observe the periodicity captured in the heatmap. The second involves high-jumps; notice that we are roughly able to align the running and jumping phases though they happen at different times. Heatmaps (right) represent relative block similarities from different time-steps of the videos. The color of the frame boxes describe the associated actions; matching colors broadly represent similar action stages.}}
  \label{same-align-two-views}
\end{figure*}

\subsection{Qualitative Results}
We motivate CoCon arguing about the benefits of preserving similarities across view-specific spaces. We observe respecting structure across views results in emergence of higher-order semantics without additional supervision e.g. sensible class relationships and good feature representations. Jointly training with views known to perform well for video action understanding allows us to learn good video representations, consequently, imparting unexpected side effects such as action alignment across videos. We discuss various experiments and results to support these claims.

\textbf{t-SNE Visualization}
We explore t-SNE visualizations of our learned representations on the $1^{st}$ test split of UCF101 extracted using $F(.)$. For clarity, only 21 action classes are displayed. We loosely order the action classes according to their relationships. Classes having similar colors are semantically similar. We can roughly observe the following broad categories present in the mentioned classes: Playing Instruments, Water Sports, Physical Sports, Physical Activities, Makeup-Hygiene. Results are displayed in Fig \ref{tsne}. Although we operate in a self-supervised setting, CoCon is able to uncover deeper semantic features allowing us to uncover inter-class relationships. We can see a much more concise and consistent clustering in CoCon compared to CPC.

\textbf{Effect of action classes on performance} 
Figure \ref{class_acc} shows the classes which observe the least and highest performance improvements when co-trained with multiple views. We observe a loose pattern where action classes involving distinguishable physical movements see larger improvements. We can argue this is because we use views which are suitable for physically intensive actions.

\textbf{Inter-Class Relationships}
In order to study consistency of structure across different views, we look at relationships between classes by inferring their similarities through our learned features. We compare cosine similarities across videos from different classes and compute the most similar four classes for each action. We repeat the process for all views and look at the consistency of the results. We only display classes which are amongst the closest ones across all views. Table \ref{table:nn_classes} summarizes our results. We see the detected nearest actions are semantically related to the original actions. In the cases of PlayingCello, we encounter a cluster of categories involving playing instruments. Similarly for BasketBall, we can see emergence of sports-based relationships even though there isn't any visual commonality between the categories. It's worth noting that as these nearest classes are consistent across different views, our approach cannot cheat to generate them i.e. it cannot look at 'background crowd' or 'green field' and infer that a video clip is related to sports. Since views such as Optical-Flow, SegMasks and KeypointHeatmap do not have such information and are very low-dimensional.

\textbf{Action Alignment}
Even though we only use self-supervision, our embeddings are able to capture relevant semantics through our multi-view approach allowing loose alignment between videos. To compute this soft alignment, we divide each video into 18 blocks and compute block-level features. We then utilize relative cosine similarities to infer associations between the videos. Figure \ref{same-align-two-views} highlights a few examples. Notice the periodicity implicitly present in some actions (e.g. pullups) captured through the heatmap allowing us to perform non-linear alignment.

\section{Conclusion}

We propose a cooperative version of contrastive learning, called CoCon, for self-supervised video representation learning. By leveraging relationships across views, we encourage our self-supervised learning objective to be aligned with the underlying semantics. We demonstrate the effectiveness of our approach on the downstream task of action classification, and illustrate the semantic structure of our representation.
We show that additional input views generated by off-the-shelf computer vision algorithms can lead to significant improvements, even though they are noisy and derived from an existing modality i.e. RGB. As these views are 'freely' available, this shows the feasibility of utilizing multi-view approaches on datasets which are not traditionally considered multi-view. We hope this enables the ability to leverage multi-view learning algorithms and observe performance gains even on single-view datasets.

\noindent\textbf{Acknowledgement.}
This work has been supported by Toyota Research Institute (TRI). This article solely reflects the opinions and conclusions of its authors and not TRI or any entity associated with TRI.


{\small
\bibliographystyle{ieee_fullname}
\bibliography{cvpr}
}

\newpage

\twocolumn[{%
\renewcommand\twocolumn[1][]{#1}%
\begin{center}
{\centering{\Large{\textbf{CoCon: Cooperative-Contrastive Learning}}}}\\
{\centering{\large{\textbf{Supplementary Material}}}}\\
\vspace{5mm}
\end{center}
}]

\section*{A. Additional Details}
\label{appendix}

\subsection*{A.1. Model Overview}
\label{model-overview}

We build our framework borrowing the learning framework present in \cite{dpc} which learns video representations through spatio-temporal contrastive losses. It should be noted that even though we use this particular self-supervised backbone in our experiments, our approach is not restricted by the choice of the underlying self-supervised task.

A video $V$ is a sequence of $T$ frames (not necessarily RGB images) with resolution $H \times W$ and $C$ channels, $\{\mathbf{i}_1, \mathbf{i}_2, \ldots, \mathbf{i}_{T}\}$, where $\mathbf{i}_t \in \mathbb{R}^{H \times W \times C}$. Assume $T = N * K$, where $N$ is the number of blocks and $K$ denotes the number of frames per block. We partition a video clip $V$ into $N$ disjoint blocks $V = \{\mathbf{x}_1, \mathbf{x}_2, \ldots, \mathbf{x}_{N}\}$, where $\mathbf{x}_j \in \mathbb{R}^{K \times H \times W \times C}$ and a non-linear encoder $f(.)$ transforms each input block $x_j$ into its latent representation $z_j = f(x_j)$. 

An aggregation function, $g(.)$ takes a sequence $\{z_1, z_2, \ldots, z_j\}$ as input and generates a context representation $c_j = g(z_1, z_2, \ldots, z_j)$. In our setup, $z_j \in \mathbb{R}^{H' \times W' \times D}$ and $c_j \in \mathbb{R}^{D}$. $D$ represents the embedding size and $H'$, $W'$ represent down-sampled resolutions as different regions in $z_j$ represent features for different spatial locations. We define $z'_j = Pool(z_j)$ where $z'_j \in \mathbb{R}^{D}$ and $c = F(V)$ where $F(.) = g(f(.))$. In our experiments, $H'=4, W'=4, D=256$.

To learn effective representations, we create a prediction task involving predicting $z$ of future blocks similar to \cite{dpc}. In the ideal scenario, the task should force our model to capture all the necessary contextual semantics in $c_t$ and all frame level semantics in $z_t$. We define $\phi(.)$ which takes as input $c_t$ and predicts the latent state of the future frames. The formulation is given in Eq. \eqref{pred-eq}.

\begin{equation}
    \begin{split}
        \widetilde{z}_{t+1} = \phi(c_t),\\
        \widetilde{z}_{t+1} = \phi(g(z_1, z_2, \ldots, z_t)),\\
        \widetilde{z}_{t+2} = \phi(g(z_1, z_2, \ldots, z_t, \widetilde{z}_{t+1})),
    \end{split}
    \label{pred-eq}
\end{equation}
where $\phi(.)$ takes $c_t$ as input and predicts the latent state of the future frames. We then utilize the predicted $\widetilde{z}_{t + 1}$ to compute $\widetilde{c}_{t+1}$. We can repeat this for as many steps as we want, in our experiments we restrict ourselves to predict till 3 steps in to the future.
 
Note that we use the predicted $\widetilde{z}_{t+1}$ while predicting $\widetilde{z}_{t+2}$ to force the model to capture long range semantics. We can repeat this for a varying number of steps, although the difficulty increases tremendously as the number of steps increases as seen in \cite{dpc}. In our experiments, we predict the next three blocks using the first five blocks.

\begin{figure}[t]
      \centering
        \includegraphics[width=\linewidth]{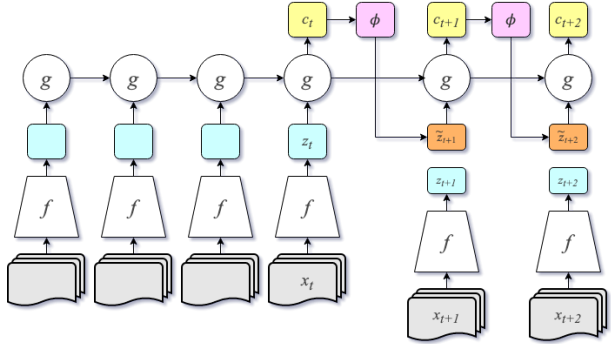}
        \caption{A diagram of the learning framework utilized. We look at features in a sequential manner while simultaneously trying to predict representations for future states.}
        \label{learningFramework}
\end{figure}

\subsection*{A.2. Datasets}

\textbf{Kinetics400} contains 400 human action classes, with at least 400  real-world video clips for each action. Each clip lasts around 10s and is taken from a different YouTube video. The actions are human focused and cover a broad range of classes including human-object and human-human interactions. The large diversity and variance in the dataset make it an extremely challenging dataset.

\textbf{HMDB51} dataset contains around 6800 real-world video clips from 51 action classes. These action classes cover a wide range of actions - facial actions, facial action with object manipulations, general body movement, and general body movements with human interactions. This dataset is challenging as it contains many poor quality video with significant camera motions and also the number of samples are not enough to effectively train a deep network. We report classification accuracy for 51 classes across 3 splits provided by the authors.

\textbf{UCF101} dataset contains 13320 videos from 101 action classes that are divided into 5 categories - human-object interaction, body-movement only, human-human interaction, playing musical instruments and sports. Action classification in this datasets is challenging owing to variations in pose, camera motion, viewpoint and spatio-temporal extent of an action.

\subsection*{A.3. Views}

We simultaneously learn encoders for RGB and Optical Flow while training on Kinetics-400. Instead of using the commonly used TVL1-Flow, we rely on Farneback flow which usually results in lower performance for action classification, however is much faster to compute. For UCF101 and HMDB51, we simultaneously learn encoders for RGB, TVL1 Optical Flow, Pose Heatmaps and Semantic Maps.

We give a brief overview of the views utilized and their generation.
\begin{itemize}
    \item \textbf{RGB Images, RGB} - We directly use sequences of RGB frames present in videos
    \item \textbf{Optical Flow, Flow} - We use the popular TVL1 flow for UCF101 and HMDB51 and Farneback Flow (FF) for Kinetics400. FF is known to perform worse than TVL1-Flow on visual recognition tasks, however, it is quicker to compute. This view mismatch views leads to lesser gains when using Kinetics pre-trained flow weights for UCF101 and HMDB51.
    \item \textbf{Pose Keypoint Heatmaps, PoseHMs} - We use an off-the-shelf keypoint detector \cite{detectron2} and extract confidence heatmaps for each keypoint. Note that we perform no pre/post-processing on the results and directly use this as input to our model. The input modality is inherently very noisy, however, we still observe improved performance. \item \textbf{Human Segmentation Masks, SegMasks} - Similar to the above, we use an off-the-shelf semantic segmentation network \cite{detectron2} and extract confidence scores for human segmentation. Similar to pose keypoint heatmaps, this input modality is inherently very noisy.
\end{itemize}

\begin{figure*}[t]
    \centering
    \includegraphics[width=\linewidth]{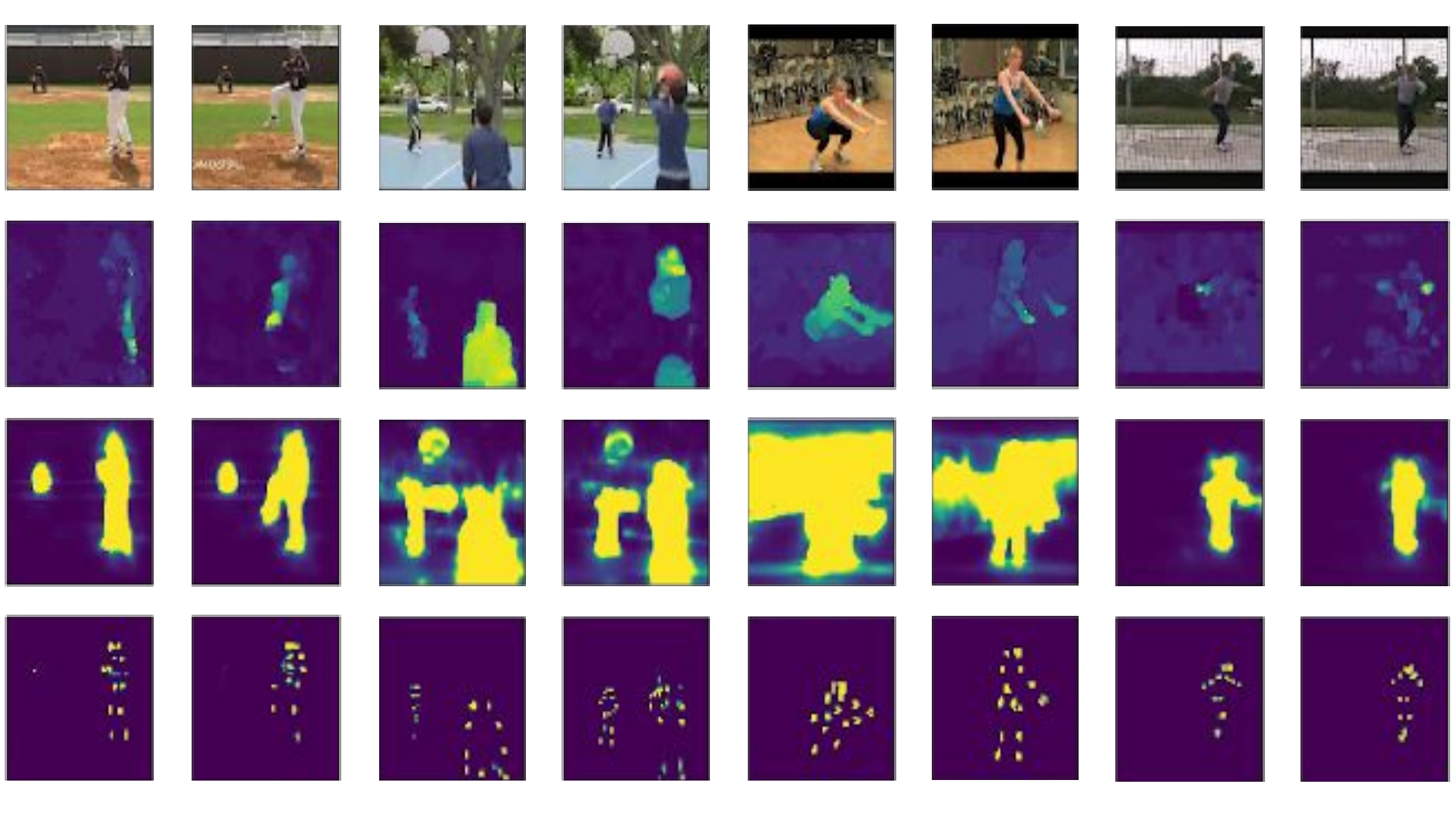}
    \captionof{figure}{Examples for each view. From top to bottom - RGB, Flow, SegMasks and Poses.}
    \label{views-samples-full}
\end{figure*}

Fig. \ref{views-samples-full} shows examples of different views. Note the prevalence of noise in a few samples, specially SegMasks. There are multiple other instances where PoseHMs are noisy as we're unable to even localize the actor accurately.

\subsection*{A.4. Implementation Details}

\label{big-impl}

We choose to use a 3D-ResNet similar to \cite{hara2018can} as the encoder $f(.)$. Following \cite{dpc} we only expand the convolutional kernels present in the last two residual blocks to be 3D ones. We used 3D-ResNet18 for our experiments, denoted as ResNet18. We use a weak aggregation function $g(.)$ in order to learn a strong encoder $f(.)$. Specifically, we use a one-layer Convolutional Gated Recurrent Unit (ConvGRU) with kernel size (1, 1) as $g(.)$. The weights are shared amongst all spatial positions in the feature map. This design allows the aggregation function to propagate features in the temporal axis. A dropout \cite{dropout} with $p = 0.1$ is used when computing the hidden state at each time step. A shallow two-layer perceptron is used as the predictive function $\phi(.)$. Recall $z_j' = Pool(z_j) where z'_j \in R_D$. We utilize stacked max pool layers as $Pool(.)$.

To construct blocks to pass to the network, we uniformly choose one out of every 3 frames. We then group these into 8 blocks containing 5 frames each. Since the videos we use are usually 30fps, each block roughly covers 0.5 seconds worth of content. The predictive task we design involves predicting the last three blocks using the first five. Therefore, we effectively predict the next 1.5 seconds based on the first 2.5 seconds.

We perform random cropping, random horizontal flipping, random greying, and color jittering to perform data augmentation in the case of images. For optical flow, we only perform random cropping on the image. As discussed earlier, Keypoint Heatmaps and Segmentation Confidence Masks are modelled as images, therefore we perform random cropping and horizontal flipping in their case. Note that random cropping and flipping is applied for the entire block in a consistent way. Random greying and color jittering are applied in a frame-wise manner to prevent the network from learning low-level features such as optical flow. Therefore, each video block may contain both colored and grey-scale image with different contrast.

All individual view-specific models are trained independently using only $\mathcal{L}_{cpc}$. After which we proceed to train all view-specific models simultaneously using $\mathcal{L}_{cocon}$. All models are trained end-to-end using Adam \cite{adam} optimizer with an initial learning rate $10^{-3}$ and weight decay $10^{-5}$. Learning rate is decayed to $10^{-4}$ when validation loss plateaus. A batch size of 16 samples per GPU is used, and our experiments use 4 GPUs. We train models on UCF101 for 100 epochs using $\mathcal{L}_{cpc}$, after which they are collectively trained together for 60 epochs using $\mathcal{L}_{cocon}$. We repeat the same for Kinetics400 with reduced epochs. We train models on Kinetics400 for 80 epochs using $\mathcal{L}_{cpc}$ and further for 40 epochs using $\mathcal{L}_{cocon}$.

The learned representations are evaluated by their performance on the downstream task of action classification. We follow the evaluation practice from recent works and use the weights learned through our self-supervised framework as initialization for supervised learning. The whole setup is then fine-tuned end-to-end using class label supervision. We finally report the fine-tuned accuracies on UCF101 and HMDB51. We use the learned composite function $F(.)$ to generate context representations for video blocks. The context feature is further passed through a spatial pooling layer followed by a fully-connected layer and a multi-way softmax for action classification. We use dropout with $p=0.7$ for classification. The models are fine-tuned for 100 epochs with learning rate decreasing at different steps. During inference, video clips from the validation set are densely sampled from an input video and cut into blocks with half-length overlapping. The softmax probabilities are averaged to give the final classification result.

\section*{B. Additional Results}

We motivate CoCon arguing about the benefits of preserving similarities across view-specific feature spaces. We observe respecting structure across views results in emergence of higher-order semantics without additional supervision e.g. sensible class relationships and good feature representations. We go over different results in the following sections.

\subsection*{B.1. t-SNE Visualization}

\begin{figure*}[h]
  \centering
      \includegraphics[width=\linewidth]{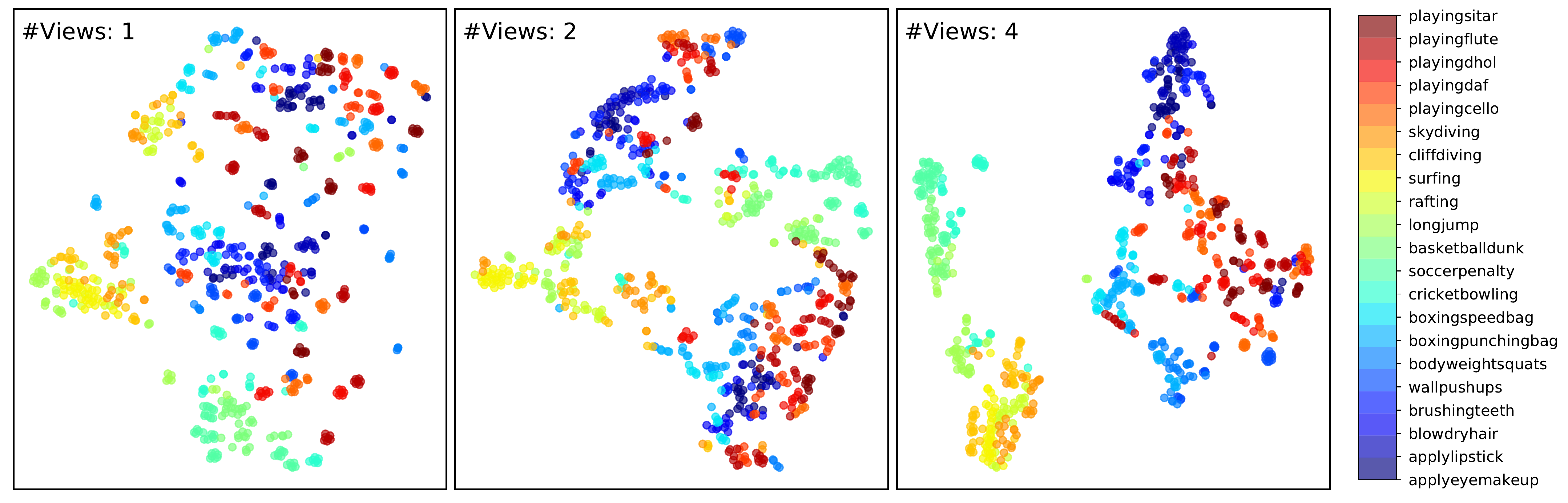}
  \caption{Emergence of relationships between different actions using CoCon with varying number of views. Note that CoCon becomes the same as CPC when $\#views=1$}
  \label{tsne-all}
\end{figure*}

We explore t-SNE visualizations of our learned representations on the $1^{st}$ test split of UCF101 extracted using $F(.)$. Our model is trained on the corresponding train split to ensure we're testing out of sample quality. For clarity, only 21 action classes are displayed. We loosely order the action classes according to their relationships. Classes having similar colors are semantically similar. Results are displayed in Fig \ref{tsne-all}. Even though we operate in a self-supervised setting, our approach is able to uncover deeper semantic features allowing us to uncover inter-class relationships. We can see a much more concise and consistent clustering in CoCon compared to CPC. We also observe the distinct improvement in the compactness of the clusters as we increase the number of views.

\subsection*{B.2. Inter-Class Relationships}

In order to study the manifold consistency across different views, we look at relationships between classes by inferring their similarities through the learned features. We compare cosine similarities across video clips from different classes. We then compute the most similar five classes for each action. We repeat the process for all views and look at the consistency of the results. Ideally, semantically similar classes should be consistent across all views, assuming the views reasonably capture the essence of the task we're interested in.

We observe that CoCon leads to much higher consistency across different views. Specifically, we see ~41 classes which have at least four out of five top-classes consistent in all views; as opposed to ~10 classes in CPC. Similar patterns are seen when we consider other thresholds. In order to confirm that the nearest classes are actually sensible, we mention the most-similar classes for a few action classes.

\begin{table*}[h]
    \renewcommand{\arraystretch}{1.0}
    \renewcommand{\tabcolsep}{2mm}
    \centering
    {\small
    \begin{tabular}{c|c|c}
        \toprule
        \multirow{2}{*}{Action Class} & \multicolumn{2}{c}{Nearest Classes} \\
        & CoCon & CPC\\
        \midrule
            skiing & surfing, skijet & surfing \\
            playingcello & playingsitar, playingtabla, playingdhol & N/A\\
            jumpingjack & jumprope, pullups, bodyweightsquats, cleanandjerk & N/A\\
            basketball & baseballpitch, cricketshot, fieldhockey, cricketbowling & N/A\\
            hammerthrow & baseballpitch, throwdiscus, shotput & N/A\\
            wallpushups & writingonboard, bodyweightsquats & N/A\\
            brushingteeth & applylipstick, applyeyemakeup, shavingbeard, haircut & applylipstick\\
        \bottomrule
    \end{tabular}
    }
    \caption{Closest semantic classes provided by different models. CPC has very few consistent nearest classes across views. While views trained using CoCon show consistent results across views, leading to sensible inter-class relationships}
\end{table*}

We can see that the nearest actions generated are semantically related to the original actions. In the cases of PlayingCello, we encounter a cluster of categories involving playing instruments. Similarly for BasketBall, we can see emergence of sports-based relationships even though there is no visual commonality between categories. We also see a few seemingly unrelated classes as well, e.g., BoxingPunchingBag and YoYo; SalsaSpin and WalkingWithDog. A deeper inspection into the samples is required to comment whether this truly makes sense. It is worth noting that as these nearest action classes are mostly consistent across different views, our approach cannot cheat to generate them i.e. it cannot look at 'background crowd' or 'green field' and infer that the video clip is related to sports. Since views such as Optical-Flow, SegMasks and KeypointHeatmap do not have such information and are much low-dimensional.

\section*{C. Action Alignment}

\begin{figure*}[h]
  \begin{subfigure}{\textwidth}
  \centering
    \includegraphics[width=\linewidth]{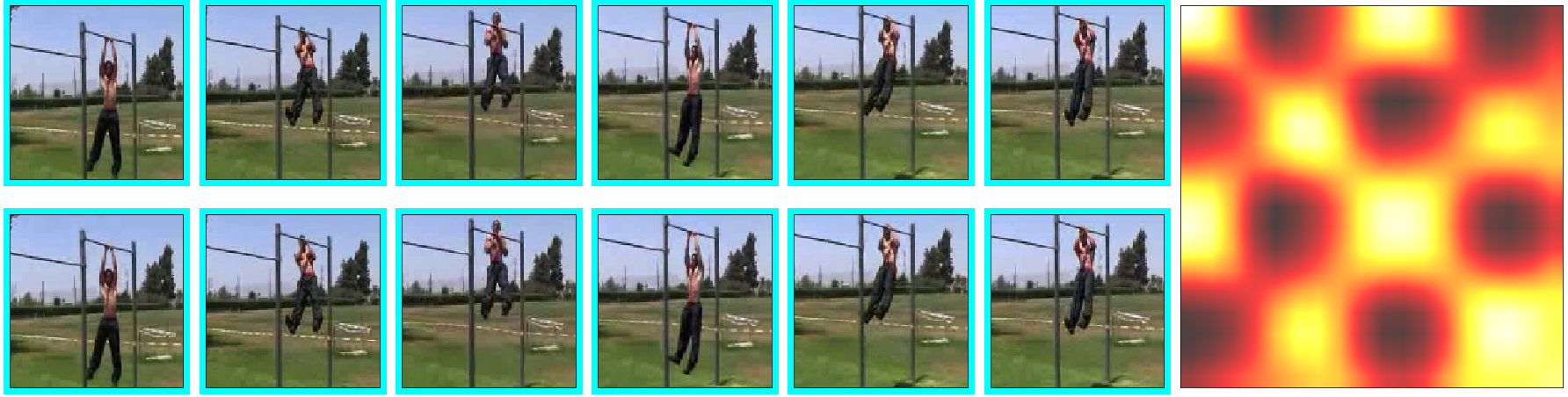}
  \end{subfigure}
   \begin{subfigure}{\textwidth}
  \centering
    \includegraphics[width=\linewidth]{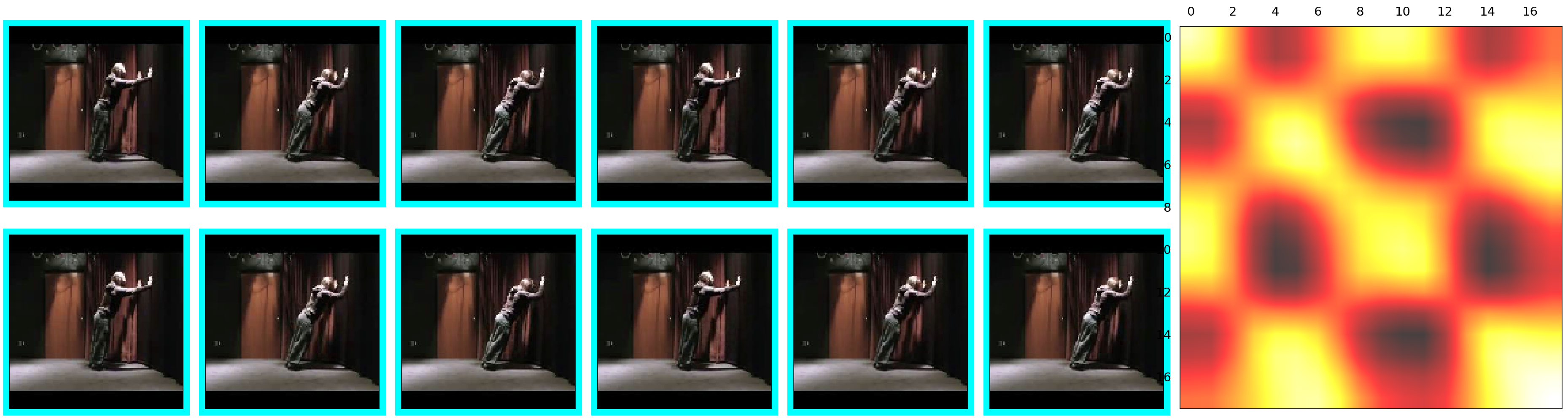}
  \end{subfigure}
  \caption{\small{Soft Alignment of actions between the same video instances. The heat-map represents the relative similarities between blocks at various timesteps. Notice periodic patterns in the actions.}}
  \label{align-same-video}
\end{figure*}

An interesting side-effect of improved representations for actions is the possibility of performing loose action alignment. Even though we only use self-supervision, CoCon embeddings are able to capture relevant semantics through our multi-view approach allowing loose alignment between videos. To compute this soft alignment, we divide each video into 18 blocks and compute block-level features $z'$. We then utilize relative cosine similarities to infer associations between the videos. We smoothen the heatmap in order to make it visually appealing. Figure \ref{same-align-two-views} shows alignment between different videos. Figure \ref{align-same-video} highlights a few examples when we perform alignment between same videos. Notice the periodicity implicitly present in these actions captured through the heatmap.

\subsection*{C.1. Cosine similarity}

This section highlights the ability of representation generated through CoCon to capture meaningful semantics going beyond low-level features. We look at cosine similarity distributions of video representation from UCF101. We extract one context representation for each video and pool it into a vector. We then compute the cosine similarity for each pair of video features across the unseen UCF101 test set. The cosine distance is summarized by a histogram, where the 'blue' histogram represents the score distribution for positives i.e. videos belonging to the same class; and the 'orange' one shows the distribution for negatives i.e. videos from different classes.

\begin{figure*}
    \centering
    \begin{subfigure}{.5\textwidth}
        \centering
        \includegraphics[width=0.95\textwidth]{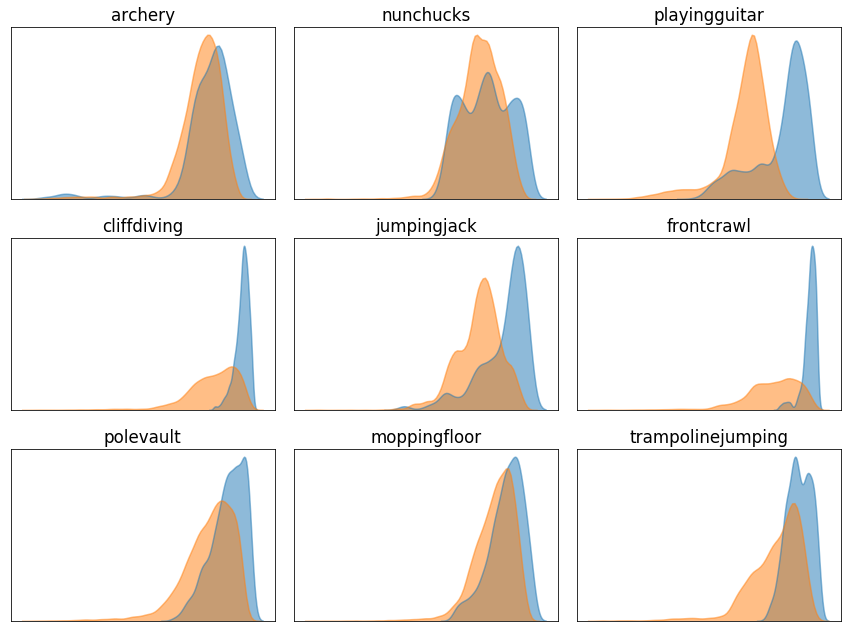}
        \caption{Generated by {CoCon}}
        \label{anti-cossim-cocon}
    \end{subfigure}%
    \begin{subfigure}{.5\textwidth}
        \centering
        \includegraphics[width=0.95\textwidth]{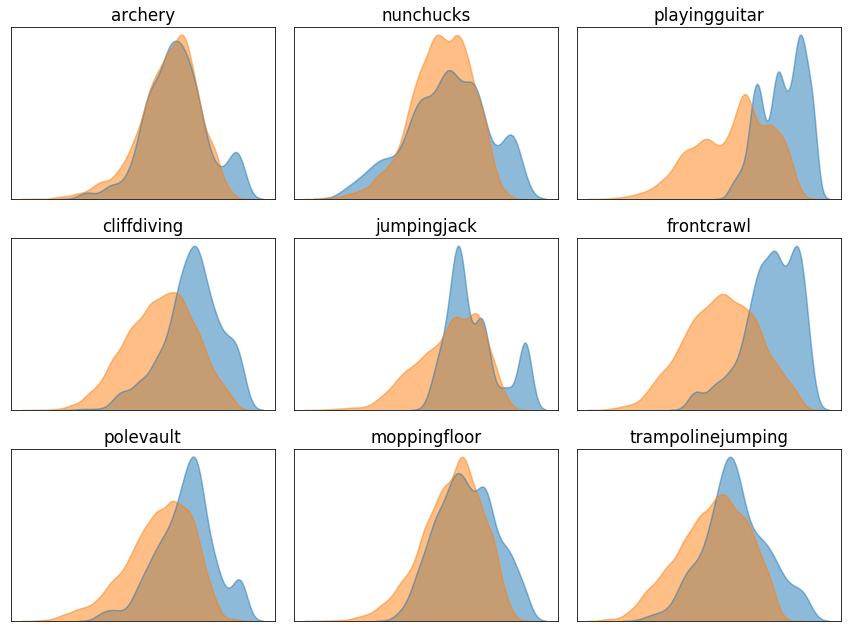}
        \caption{Generated by CPC}
        \label{anti-cossim-cpc}
    \end{subfigure}
    \caption{Distributions of cosine-similarity scores between representations of videos from the same (blue) and other classes (red).}
    \label{cossim}
\end{figure*}

\subsection*{C.2. Nearest Neighbors}

We utilize CoCon to perform video retrieval for different query videos. Note that CoCon is able to look past purely visual background features and focus on actions even though it only used RGB inputs. For example, we see that we are able to retrieve close neighbors for BenchPress, even though it is very visually different with varying poses. For the IceDancing sample, even though it incorrectly considers onbe video where the person is running, we can still see similarities between the underlying actions in the videos. Similar results can be seen in other examples as well. This hints towards the fact that CoCon representation are able to capture action semantics even while using RGB views.

\begin{figure*}
    \centering
    \label{nn-imgs}
    \includegraphics[width=0.78\linewidth]{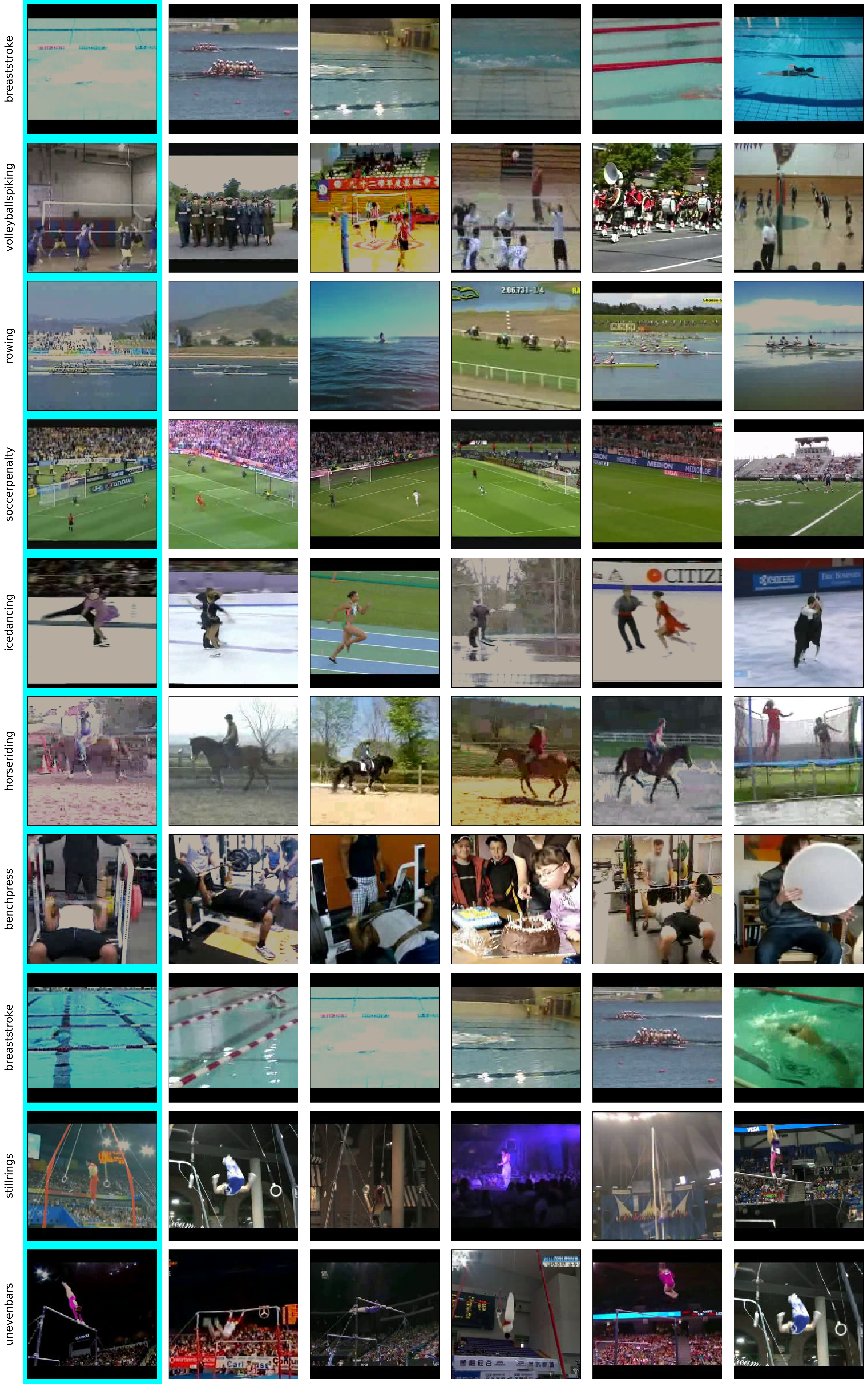}
    \caption{Nearest neighbors computed using RGB representations. Query video is highlighted on the left with \textcolor{aqua}{Aqua Blue}.}
\end{figure*}

\end{document}